\documentclass[conference]{IEEEtran}
\IEEEoverridecommandlockouts
\usepackage{cite}
\usepackage{amsmath,amssymb,amsfonts}
\usepackage{algorithmic}
\usepackage{graphicx}
\usepackage{textcomp}
\usepackage{xcolor}
\usepackage[most]{tcolorbox}
\usepackage{hyperref}
\usepackage{booktabs}
\usepackage[table]{xcolor}
\RequirePackage{url}
\RequirePackage[svgnames,dvipsnames]{xcolor}
\definecolor{tpurple}{HTML}{574389}
\definecolor{citec}{HTML}{03ff03}
\definecolor{urlc}{HTML}{ed058d}

\hypersetup{
    colorlinks=true,
    linkcolor=urlc,
    citecolor=citec,
    urlcolor=urlc
}
\definecolor{best_green}{HTML}{DDF4DD}
\definecolor{second_purple}{HTML}{E4E0FF}
\definecolor{deltapurple}{RGB}{81,45,168}
\definecolor{deltagreen}{RGB}{0,100,60}
\definecolor{refgray}{gray}{0.45}
\definecolor{findingbg}{HTML}{E8E9F2}
\definecolor{findingpurple}{HTML}{584489}

\newcommand{\best}[1]{\cellcolor{best_green}\textbf{#1}}
\newcommand{\second}[1]{\cellcolor{second_purple}\underline{#1}}

\newcommand{\besttext}[1]{%
  {\setlength{\fboxsep}{0.6pt}\colorbox{best_green}{#1}}%
}
\newcommand{\secondtext}[1]{%
  {\setlength{\fboxsep}{0.6pt}\colorbox{second_purple}{#1}}%
}
\definecolor{findingbg}{HTML}{E8E9F2}
\definecolor{findingpurple}{HTML}{584489}
\newtcolorbox{findingbox}{
  colback=findingbg,
  colframe=findingpurple,
  boxrule=0.5pt,
  arc=1.5pt,
  left=2pt,
  right=2pt,
  top=3pt,
  bottom=2pt,
  before skip=6pt,
  after skip=2pt
}

\def\BibTeX{{\rm B\kern-.05em{\sc i\kern-.025em b}\kern-.08em
    T\kern-.1667em\lower.7ex\hbox{E}\kern-.125emX}}
\begin{document}

% \title{Seeing More or Being Told More? On the Limits of Test-Time Scaling in GUI Grounding}
% {\footnotesize \textsuperscript{*}Note: Sub-titles are not captured for https://ieeexplore.ieee.org  and

\title{%
\texorpdfstring{%
  \raisebox{-0.20em}{%
    \includegraphics[
      height=1.25em,
      trim=17 14 0 9,
      clip
    ]{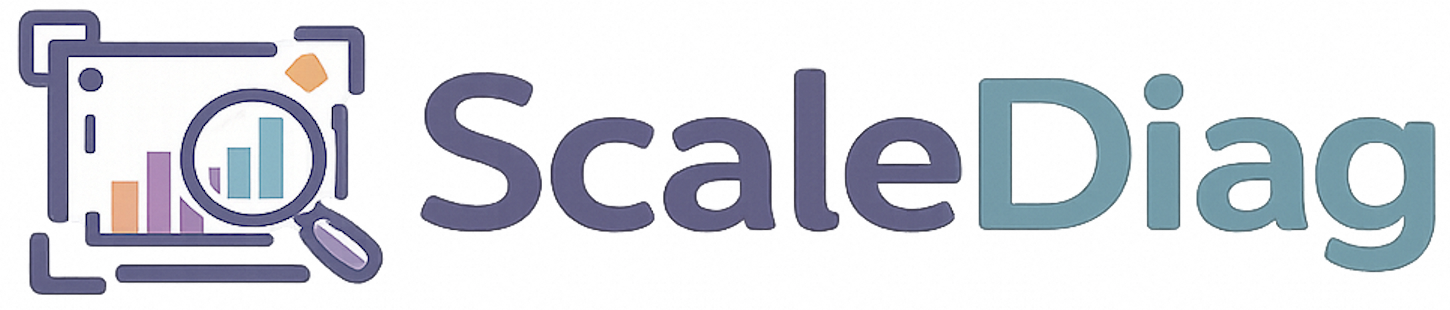}%
  }%
  \kern-0.01em:\kern0.12em%
  \textnormal{Diagnosing the Limits of Test-Time Scaling in GUI Grounding}%
}{%
  ScaleDiag: Diagnosing the Limits of Test-Time Scaling in GUI Grounding%
}%
}

% \author{
% \IEEEauthorblockN{Anonymous Authors}
% \IEEEauthorblockA{Paper under double-blind review}
% }
\author{
\IEEEauthorblockN{
Chen Liu\textsuperscript{\dag}\quad
Zhen Fang\textsuperscript{\dag}\quad
Xin Yu\textsuperscript{\ddag}\quad
Dadong Wang\textsuperscript{\S}\quad
Ling Chen\textsuperscript{\dag *}\thanks{
*Corresponding author: Ling Chen (ling.chen@uts.edu.au).
}
}
\IEEEauthorblockA{
\small
\textsuperscript{\dag}\textit{University of Technology Sydney, Australia},
{\footnotesize\texttt{\{chen.liu, fang.zhen, ling.chen\}@uts.edu.au}}\\
\textsuperscript{\ddag}\textit{Adelaide University, Australia},
{\footnotesize\texttt{xin.yu@adelaide.edu.au}}
\quad
\textsuperscript{\S}\textit{CSIRO Technology, Australia},
{\footnotesize\texttt{Dadong.Wang@csiro.au}}
}
}

\maketitle

\begin{abstract}
GUI grounding maps a screenshot and a natural language instruction to the coordinates of the intended interface element. 
Existing approaches improve grounding through \textit{visual scaling}, \textit{instruction scaling}, and \textit{sampling scaling}, yet it remains unclear which grounding bottlenecks these strategies address and where their gains saturate.
To address this gap, we introduce \textbf{ScaleDiag}, a controlled diagnostic framework that isolates the effects of visual input, instruction information, and prediction sampling on GUI grounding by independently varying target scale and surrounding context, target-specific information and auxiliary descriptions, and candidate coverage and selection, respectively.
We evaluate ScaleDiag across five benchmarks and up to eight vision-language models and find that each scaling strategy operates within a much narrower regime than commonly assumed. 
(1) Visual scaling is governed by target occupancy rather than a fixed crop ratio, with full-screen errors typically rescued once the target occupies only 0.32\% of the cropped view, while further zooming eventually becomes harmful. 
(2) Instruction scaling is driven by information that identifies the target rather than description richness itself, as even ground-truth-derived attributes such as color, size, and shape are often neutral or harmful, while appearance and spatial descriptions account for over 90\% of the achievable gain. 
(3) Repeated sampling exposes a different limitation, as correct coordinates are frequently generated but remain difficult to select, leaving 5.1--11.0 points of oracle headroom while geometric consensus provides almost no improvement. Together, these results show that test-time scaling is constrained by distinct operating regimes rather than benefiting monotonically from more visual focus, richer instructions, or additional samples.
Our code is available at \url{https://github.com/YenanLiu/ScaleDiag}.

\vspace{0.5em}

\end{abstract}

\vspace{-1.0em}

\section{Introduction}
GUI grounding maps a screenshot and a natural language instruction to the coordinates of the intended interface element, enabling GUI agents to translate high-level goals into concrete actions \cite{cheng2024seeclick, gou2025navigating, wu2410atlas}. 
Recent vision-language models have achieved strong performance on standard grounding benchmarks \cite{xie2026scaling, chen2026ui, you2024ferret, li2025ferret}, yet their predictions remain sensitive to small targets, dense interfaces, irrelevant visual content, and ambiguous instructions. 
Existing work improves grounding through local cropping \cite{luo2025visual, lee2025reguide, wu2025dimo, du2026test, lei2025textsc}, richer target descriptions \cite{chen2026ui, liu2025ui, yang2025aria, gou2025navigating}, and repeated sampling \cite{xie2026scaling, li2025screenspot, liu2026drs}.
However, these strategies are often applied uniformly across inputs or combine multiple changes at once, making it difficult to determine when additional computation is actually useful, what drives the improvement, and where the gain saturates or reverses.

\begin{figure}[tp]
    \centering
    \includegraphics[width=0.8\linewidth]{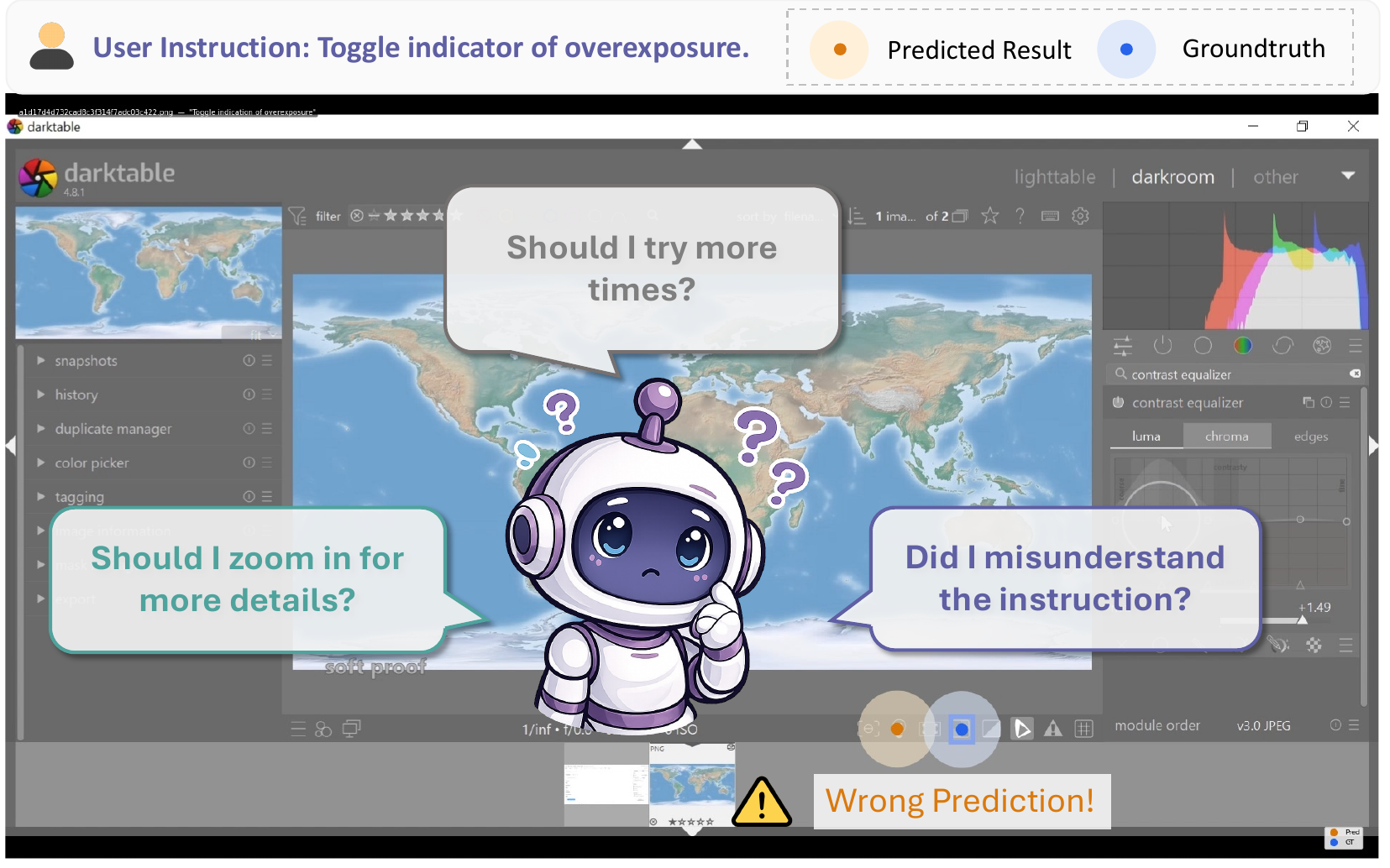}
    \vspace{-1.em}
    \caption{
    GUI grounding failures can arise from insufficient target visibility, missing target-identifying information, or unreliable candidate selection. These bottlenecks require different interventions, including visual scaling, instruction refinement, and verifier-guided sampling.
    }
    \vspace{-2.0em}
    \label{fig:motivation}
\end{figure}

To address this gap, we introduce \textbf{ScaleDiag}, a controlled diagnostic framework that isolates the effects of visual input, instruction information, and prediction sampling on GUI grounding by independently varying target scale and surrounding context, target-specific and auxiliary descriptions, and candidate coverage and selection, respectively. 
The model size is treated as a cross-cutting factor and examine whether these effects persist across models of different scales and specialization.
We evaluate ScaleDiag across five GUI grounding benchmarks and up to eight vision-language models spanning different scales and levels of GUI specialization. 
For visual scaling, we vary target-centered crop size and irrelevant context. 
For instruction scaling, we vary how much information the instruction provides to identify the target and what additional target descriptions are included.
For sampling scaling, we test whether repeated sampling generates a correct coordinate and whether that coordinate can be reliably selected from the sampled candidates.

Our experiments reveal where each form of test-time scaling helps and where its benefit breaks down.
\textit{\textcolor{tpurple}{First}}, visual scaling is governed by target occupancy rather than by a universal crop ratio. 
Full-screen errors rescued by cropping require a median target occupancy of only 0.32\%, while further enlargement eventually becomes harmful as useful context is removed. Appropriate cropping rescues 24.1\% of full-screen errors while harming only 1.3\% of correct predictions.
\textit{\textcolor{tpurple}{Second}}, instruction scaling helps when added information makes the target easier to identify, rather than simply making the description richer. 
Removing target-identifying information reduces accuracy to 4\%--6\%, while even ground-truth-derived attributes such as color, size, and shape provide little consistent benefit. 
Appearance and spatial descriptions account for over 90\% of the achievable gain, with function and goal descriptions contributing little further improvement.
\textit{\textcolor{tpurple}{Third}}, repeated sampling improves the chance of generating a correct coordinate but not of selecting it. 
Oracle@$8$ improves accuracy by 5.1--11.0 points, while coordinate-medoid selection provides almost no gain. 
Correct candidates often appear only once or twice among eight samples, making them difficult to recover through geometric consensus.
Together, these results show that visual focus, richer descriptions, and additional samples have distinct limits rather than providing monotonic gains.
In summary, our contributions are threefold:
\begin{itemize}
    \item \textbf{A controlled diagnostic framework.}
    ScaleDiag independently varies visual input, instruction information, and prediction sampling to separate effects that are coupled in existing scaling strategies.

    \item \textbf{A quantitative characterization of scaling limits.}
    We identify the target-occupancy range in which cropping helps, which instruction information contributes meaningful gains, and why additional sampled candidates fail to translate into better final predictions.

    \item \textbf{Evidence for selective test-time scaling.}
    Our results show that visual focus, richer descriptions, and additional samples are useful under different conditions, motivating test-time computation that is allocated according to the source of grounding error rather than applied uniformly.
\end{itemize}
\vspace{-0.6em}

\section{Related Work}

\begin{figure*}
    \centering
    \includegraphics[width=0.85\linewidth]{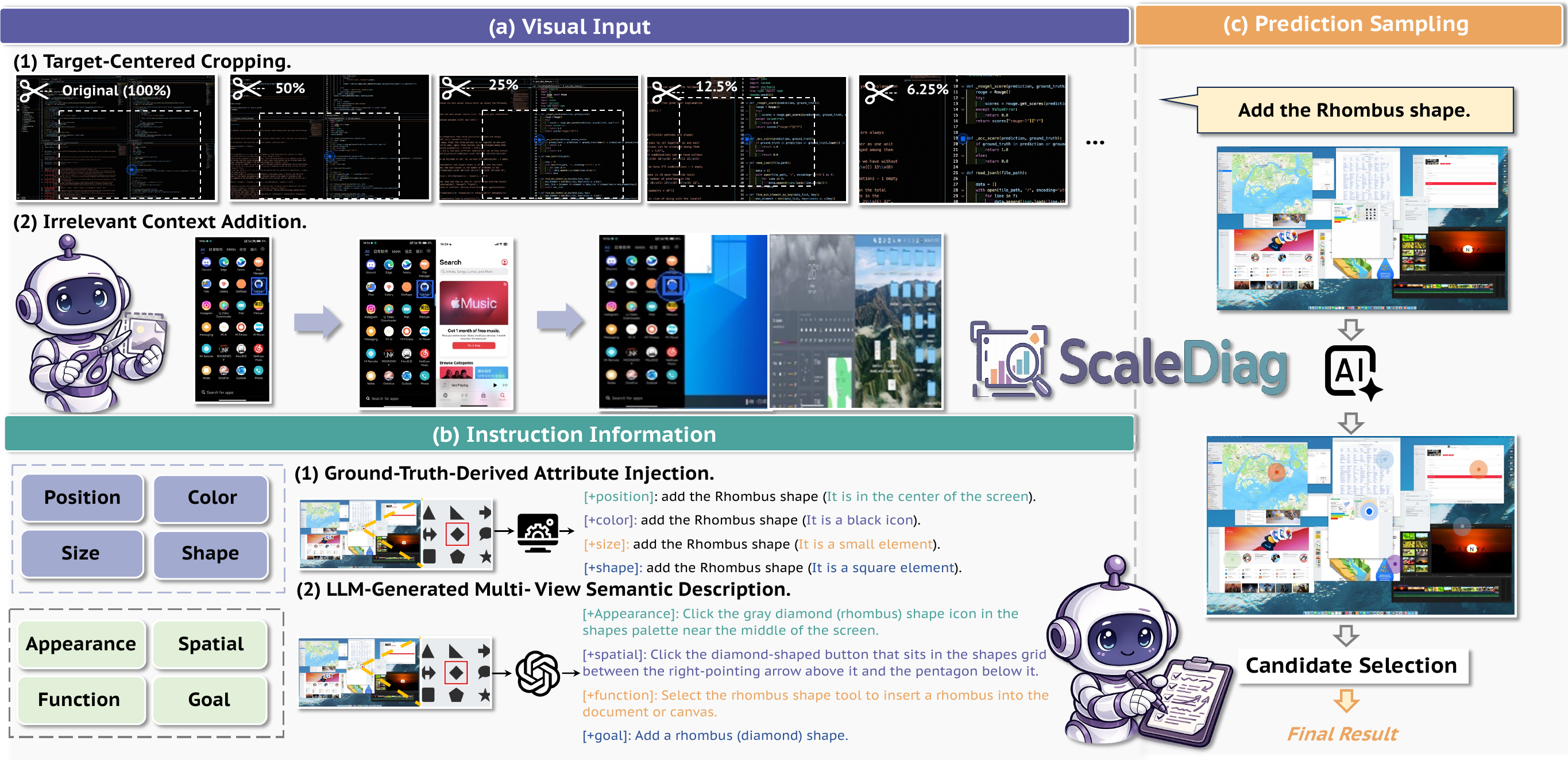}
    \vspace{-1.0em}
\caption{
Overview of our controlled diagnostic framework for GUI grounding.
(a) Controlled visual interventions vary the visual evidence while keeping the instruction fixed. We construct target-centered, aspect-ratio-preserving crops at different area ratios, and separately add resized distractor screenshots while preserving the target screenshot at its native resolution.
(b) Controlled instruction interventions keep the screenshot and target fixed while modifying the instruction. Rule-based attribute injection appends ground-truth-derived position, color, size, or shape cues, whereas an LLM generates natural descriptions from appearance, spatial, functional, and goal-oriented views.
(c) Repeated sampling keeps both the screenshot and instruction fixed and generates multiple coordinate predictions, allowing us to compare greedy decoding, coordinate-medoid consensus, and oracle candidate coverage.
Together, these interventions isolate the effects of visual evidence, instruction information, and candidate selection.
}
    \label{fig:framework}
    \vspace{-1.5em}
\end{figure*}

\subsection{Instruction Semantic Enhancement}

Successful GUI grounding requires instructions that provide enough information to identify the intended interface element \cite{cheng2024seeclick, gou2025navigating, wu2410atlas}. 
Early methods such as SeeClick \cite{cheng2024seeclick}, UGround \cite{gou2025navigating}, and OS-Atlas \cite{wu2410atlas} learn mappings from language to GUI elements using visible text, control attributes, spatial relations, and template-based referring expressions. 
These formulations work well when the target can be identified from explicit interface cues, but provide limited support for instructions expressed through appearance, function, or user intent.
More recent methods enrich the instruction with additional target descriptions \cite{yang2025aria, liu2025ui, chen2026ui}. 
Aria-UI \cite{yang2025aria}, UI-E2I-Synth \cite{liu2025ui}, and UI-Ins \cite{chen2026ui} incorporate cues such as appearance, location, function, and interaction intent, while UI-Ins further reformulates high-level intent into a target description before coordinate prediction. 
Other approaches, including MEGA-GUI \cite{kwak2025mega}, ZoomUI \cite{liu2026zoom}, and AdaZoom-GUI \cite{pei2026adazoom}, combine richer instructions with visual search to narrow the target region.
Despite these advances, it remains unclear which additional descriptions actually help identify the target and which merely add target-related information without improving grounding. 
Existing work also does not systematically test how these effects change with model scale or whether larger models can compensate when target-identifying information is missing.

\subsection{Visual Perception and Representation Enhancement}
Accurate GUI grounding requires sufficient visual detail to distinguish the target while preserving the surrounding context needed for localization \cite{hong2024cogagent, you2024ferret, li2025ferret, lin2025showui, xu2025attention}. 
Early methods improve small-target grounding through higher-resolution or multi-view encoding. CogAgent \cite{hong2024cogagent} combines global and local visual representations, while Ferret-UI \cite{li2025ferret, you2024ferret} adapts image partitioning to GUI layouts to better preserve small icons and text.
Later work reduces visual redundancy or focuses computation on task-relevant regions. 
ShowUI \cite{lin2025showui} and FocusUI \cite{ouyang2026focusui} selectively retain informative visual tokens or patches, while TAG \cite{ma2026trifuse}, GUI-AIMA \cite{zhou2025gui}, GUI-Actor \cite{wu2026gui}, and InnerZoom \cite{liu2026one} strengthen attention or computation around likely target regions.
However, visual focusing typically changes target scale and surrounding context at the same time. 
As a result, it remains unclear whether gains come from enlarging the target or removing context, how much enlargement is actually needed, and when further cropping becomes harmful.

\subsection{Test-Time Search and Verification for GUI Grounding}
When a single grounding prediction is unreliable, additional test-time computation can be used to search for, refine, or verify the target location. 
A common strategy is coarse-to-fine visual search. RegionFocus \cite{luo2025visual}, DiMo-GUI \cite{wu2025dimo}, and GUI-Eyes \cite{chen2026gui} first localize a coarse region and then refine it through cropping, zooming, or local views. 
More adaptive methods decide when and where to spend this extra computation. 
AdaZoom-GUI \cite{pei2026adazoom} selectively activates refinement for small targets, while UI-Zoomer \cite{tang2026ui} uses prediction consistency and spatial dispersion to guide zooming. DRS-GUI \cite{liu2026drs} and ZoomUI \cite{li2026gui} further formulate grounding as an iterative search process that updates the candidate region based on previous predictions or target cues.
Additional computation can also be used without explicit visual search. 
MVP \cite{zhang2026mvp} aggregates predictions from perturbed views to identify stable locations, while Re-Prefill \cite{lin2026happens} performs a second inference pass using selected visual and instruction features. These methods show that repeated inference can expose alternative predictions or reconsider an initial decision. 
However, prior work rarely distinguishes whether the gain comes from generating a correct candidate or from selecting it once generated, making it unclear how much repeated computation remains useful when candidate selection itself is the limiting factor.

\section{Method}

\subsection{Task Definition and Framework Overview}

\noindent\textbf{Task Definition.}
Given a GUI screenshot $I$ and a natural-language instruction $q$, a GUI grounding model predicts the click coordinate of the target element, $\hat{p}=(\hat{x}, \hat{y})$. 
A prediction is correct if $\hat{p}$ lies within the ground-truth bounding box $B$.

\noindent\textbf{Framework Overview.} We introduce \textbf{ScaleDiag}, a model-agnostic diagnostic framework for isolating the effects of visual input, instruction information, and prediction sampling on GUI grounding. 
ScaleDiag follows a one-factor-at-a-time design that changes one factor while keeping the target and evaluation setting fixed.
As shown in Figure~\ref{fig:framework}, the three components are studied through controlled experiments. For visual input, we vary target scale and surrounding context through target-centered cropping and added irrelevant interface content. 
For instruction information, we vary the amount of target-identifying information and the type of additional target descriptions using rule-based attributes, an information ladder, and multi-view descriptions. 
For prediction sampling, we generate multiple coordinate candidates and compare their coverage with different selection strategies to separate candidate generation from candidate selection.

\subsection{Controlled Visual Interventions}
The visual experiments keep the instruction and target fixed while varying target scale and surrounding context. 
We use target-centered cropping to change target scale while progressively reducing surrounding context, and tiled perturbations to add irrelevant content while preserving the original target.

\noindent\textbf{Target-Centered Cropping.}
To measure the effect of visual scaling without confounding it with crop-selection accuracy, we construct oracle crops using the ground-truth target location. 
Given a screenshot of size $W \times H$ and the center $c=(c_x,c_y)$ of the ground-truth bounding box, a crop with area ratio $f$ has width $W_f=\sqrt{f}W$ and height $H_f=\sqrt{f}H$.
The crop is centered at $c$ and shifted inward when necessary to remain within the screenshot while preserving its size. We evaluate $f \in \{1, 0.5, 0.25, 0.125, 0.0625, 0.03125\}$, corresponding to 100\%, 50\%, 25\%, 12.5\%, 6.25\%, and 3.125\% of the original screenshot area. 
This controlled experiment removes crop-selection errors and measures how grounding changes as the target occupies a larger fraction of the input while surrounding context is progressively reduced.

\noindent\textbf{Irrelevant Context Addition.}
Target-centered cropping simultaneously enlarges the target relative to the input and removes surrounding content. 
To isolate the effect of surrounding context, we construct tiled screenshots that keep the target screenshot unchanged while adding unrelated interfaces around it. 
On ScreenSpot-v2 \cite{screenspotv2-cheng2024seeclick}, each grid cell has the original screenshot size $W \times H$, and we construct canvases with $K \in {1,2,4}$ cells arranged as $1\times1$, $1\times2$, and $2\times2$ grids. 
The target screenshot is placed without resizing in a randomly selected cell, while the remaining cells contain screenshots sampled from the same dataset and resized to $W \times H$. 
We use a fixed random seed, $\texttt{seed}=0$, so that all models receive the same target placement and surrounding content.
Let $(o_x,o_y)$ denote the top-left offset of the target cell. 
Because the target screenshot is not resized, its bounding box $B=(x_1,y_1,x_2,y_2)$ is translated to $B'=(x_1+o_x,y_1+o_y,x_2+o_x,y_2+o_y)$.
The instruction remains unchanged. 
Thus, the target itself and its scale are preserved while only the amount of irrelevant surrounding content increases, allowing us to measure its effect on grounding independently.

\subsection{Controlled Instruction Construction}
As depicted in Figure~\ref{fig:framework} (b), all instruction experiments keep the screenshot, target element, and visual processing fixed while changing only the instruction. 
We design three controlled experiments to examine whether coarse target attributes are useful, how grounding changes with the amount of information available to identify the target, and which additional target descriptions provide further gains. 
These questions are studied through ground-truth-derived attribute injection, an instruction information ladder, and LLM-generated multi-view descriptions, respectively.

\begin{figure*}
    \centering
    \includegraphics[width=0.75\linewidth]{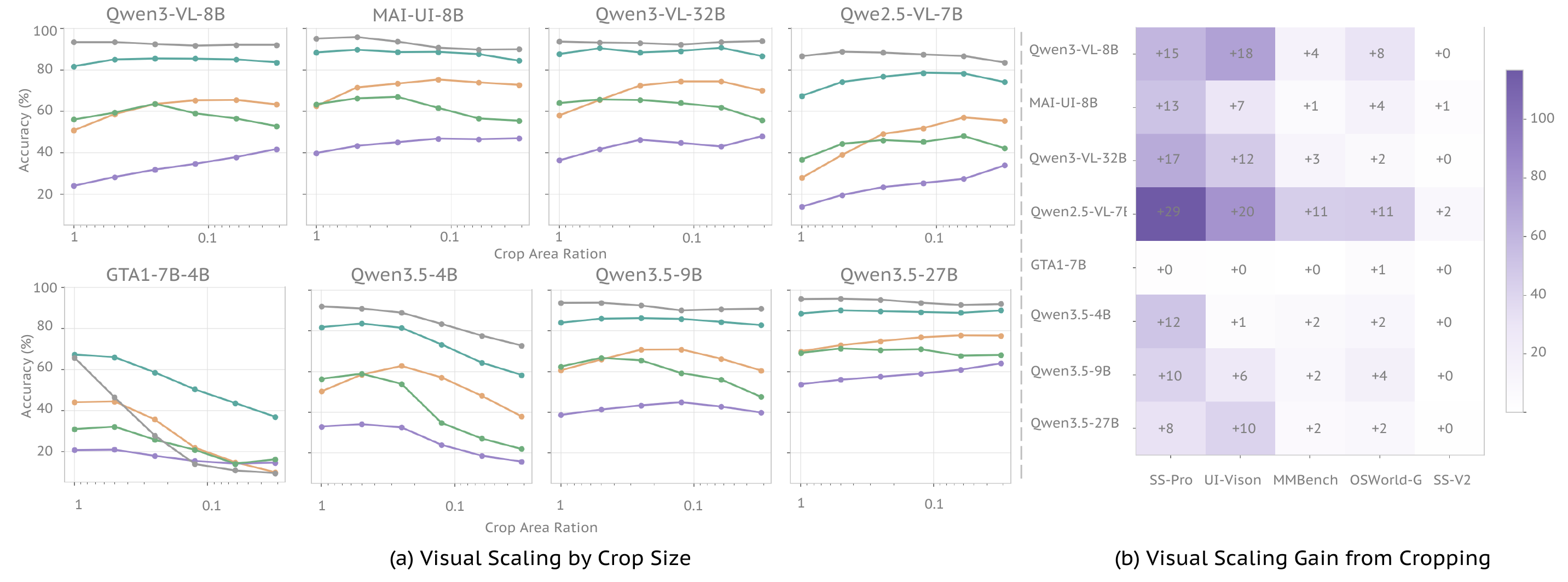}
    \vspace{-1.0em}
\caption{
Effects of target-centered cropping on GUI grounding performance.
(a) Grounding accuracy across crop area ratios for each model and benchmark.
(b) Accuracy gain of the best crop over the full screenshot.
}
    \label{fig:vis_scaling_bar}
    \vspace{-1.5em}
\end{figure*}

\noindent\textbf{Ground-Truth-Derived Attribute Injection.}
We derive four coarse attributes from the ground-truth target, including position, color, size, and shape. Let the screenshot have width $W$ and height $H$, and let $B=(x_1,y_1,x_2,y_2)$ denote the target bounding box with center $(c_x,c_y)$. 
Position is obtained by mapping $(c_x/W,c_y/H)$ to a $3\times3$ grid with left, center, and right horizontal regions and top, middle, and bottom vertical regions, producing labels such as \texttt{top-left} and \texttt{bottom-right}.
For color, we crop the target region, resize it to $16\times16$, compute the median RGB value, and assign the nearest color from a fixed vocabulary of black, white, gray, red, green, blue, yellow, orange, purple, cyan, pink, and brown using Euclidean distance in RGB space. Target size is determined by the area ratio
$a=\frac{(x_2-x_1)(y_2-y_1)}{WH}$ and labeled small when $a<5\times10^{-4}$, medium when $5\times10^{-4}\leq a<5\times10^{-3}$, and large otherwise. Shape is determined by the aspect ratio
$r=\frac{x_2-x_1}{y_2-y_1}$ and labeled wide when $r>1.6$, tall when $r<0.62$, and square otherwise.

Attributes are appended to the original instruction using the fixed template
\texttt{\textless instruction\textgreater{} (It is \textless attribute phrase\textgreater{}.)}.
For example, an attribute phrase may be
\texttt{a small blue element located in the top-right of the screen}.
We evaluate each attribute individually and four predefined combinations,
${\text{position+color},\text{position+size},\text{position+color+size},\text{all attributes}}$.
Together with the unchanged original instruction, this yields nine conditions per sample. Because all attributes are derived deterministically from ground truth, this experiment measures whether coarse target descriptions improve grounding without introducing variation from instruction generation.

\noindent\textbf{Instruction Information Ladder.}
To measure how grounding depends on the information available to identify the target, we construct a controlled ladder from an underspecified instruction to increasingly explicit target information. 
The \texttt{generic} condition removes target-identifying information and replaces the instruction with
\texttt{Click the correct interactive UI element on the screen for the task.}
The \texttt{keep-half} condition retains the first $\lceil n/2 \rceil$ tokens of an instruction with $n$ tokens, while \texttt{base} uses the complete original instruction. Starting from \texttt{base}, \texttt{clarify-pos} adds the target's coarse location on the same $3\times3$ grid, and \texttt{clarify-fine} adds its approximate horizontal and vertical positions as percentages of the screenshot dimensions. A fine-grained clarification takes the form
\texttt{It is located approximately 53\% from the left and 42\% from the top of the screen.}
This ladder measures how performance changes as target-identifying information is removed, restored, and further clarified.

\noindent\textbf{LLM-Generated Multi-View Descriptions.}
The previous experiments control coarse attributes and the amount of target-identifying information. 
We additionally study whether richer target descriptions provide further gains. Following the multi-view taxonomy of UI-Ins \cite{chen2026ui}, we use GPT-5.2 \cite{singh2025openai} to generate descriptions from four perspectives. 
The appearance view \texttt{app} describes visible text, iconography, color, and coarse location. 
The spatial view \texttt{spa} describes the target relative to neighboring interface elements. 
The functional view \texttt{fun} describes the effect of activating the target without relying on its appearance or location. The goal view \texttt{goa} describes the higher-level user intent associated with the target.

For each target, GPT-5.2 receives the original instruction, a full screenshot with the target highlighted, an enlarged target crop, and automatically derived hints for color, position, size, shape, and UI type. 
The model is instructed to follow the images when the heuristic hints conflict with the visual evidence. 
These marked images are used only to generate descriptions.
We generate four single-view descriptions, six pairwise combinations, four three-view combinations, and one four-view combination. 
Together with the original instruction, this yields 16 conditions per sample. 
Each condition integrates its selected perspectives into a single coherent instruction. Generated descriptions must remain faithful to the target, remain usable without the marked images, and avoid references to annotations, coordinates, percentages, or ground truth. 
Comparing individual views and their combinations measures which additional target descriptions improve grounding and whether their gains saturate as more descriptions are added.

\begin{figure*}
    \centering
    \includegraphics[width=0.8\linewidth]{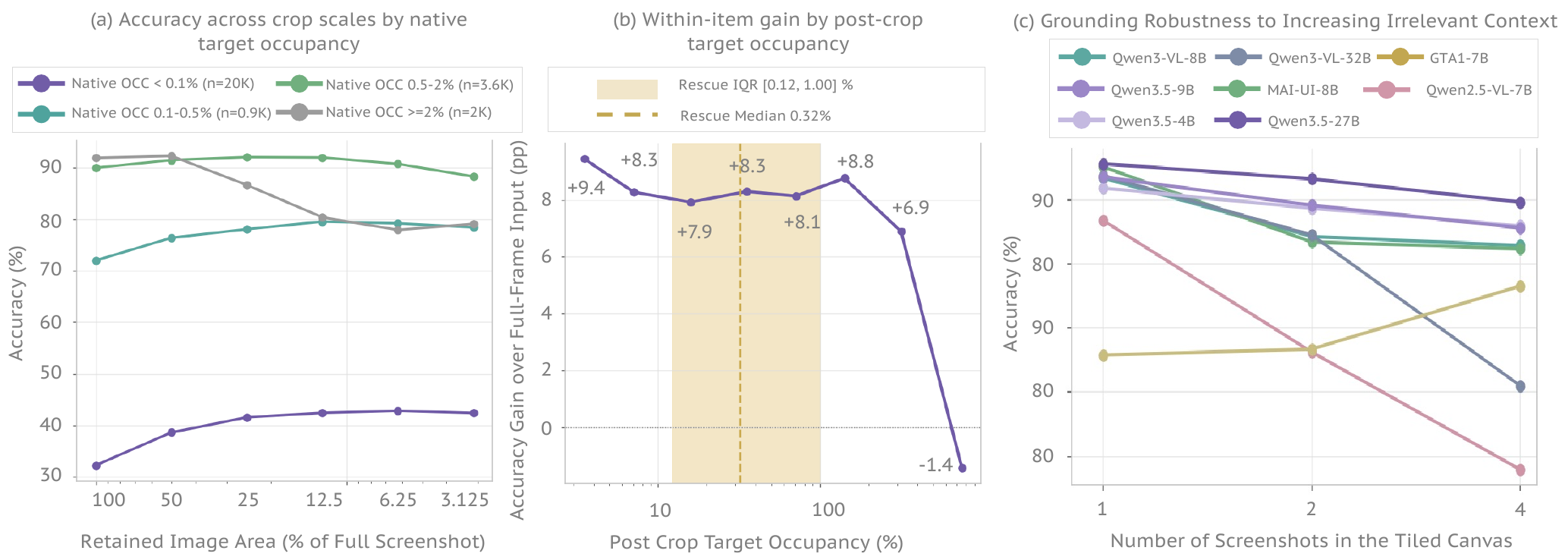}
    \vspace{-1.em}
\caption{
Difficulty-gated visual scaling under target-centered cropping.
OCC denotes the ratio of the target bounding-box area to the visible image area.
\textbf{(a)} Grounding accuracy across retained crop areas, grouped by native OCC measured on the full screenshot.
Each point reports the accuracy of one native-OCC group at a given crop scale.
\textbf{(b)} Within-item accuracy gain over the full-frame input as a function of post-crop OCC.
Each crop prediction is compared with the full-frame prediction for the same item, and the resulting changes are averaged within OCC bins.
Rescue OCC is the post-crop target occupancy at the least aggressive crop that converts a full-frame miss into a hit.
The shaded region shows its interquartile range across rescued items, while the dashed line marks its median.
Both panels pool approximately 34.9k model--benchmark--item instances.
}
    \vspace{-1.5em}
    \label{fig:visual_gated}
\end{figure*}

\begin{table*}[!t]
    \centering
    \scriptsize
    \setlength{\tabcolsep}{6.5pt}
    \renewcommand{\arraystretch}{1.0}
    \caption{
    Overall grounding accuracy and target-type breakdown across GUI grounding benchmarks.
    All denotes overall accuracy, T/I denotes text and icon accuracy, and Gap denotes their difference.
    Best and second-best results for All and Gap are highlighted in
    \besttext{green} and \secondtext{purple}, respectively.
    }
    \label{tab:baseline_type_breakdown}
    \vspace{-0.5em}
    \resizebox{0.98\textwidth}{!}{
    \begin{tabular}{lcccccccccccccc}
        \toprule
        \textbf{Method}
        & \multicolumn{3}{c}{\textbf{SS-Pro}}
        & \multicolumn{3}{c}{\textbf{UI-Vision}}
        & \multicolumn{3}{c}{\textbf{MMBench}}
        & \textbf{OSW-G}
        & \multicolumn{3}{c}{\textbf{SS-v2}} \\
        \cmidrule(lr){2-4}
        \cmidrule(lr){5-7}
        \cmidrule(lr){8-10}
        \cmidrule(lr){12-14}
        & \textbf{All}
        & \textbf{T/I}
        & \textbf{Gap}
        & \textbf{All}
        & \textbf{T/I}
        & \textbf{Gap}
        & \textbf{All}
        & \textbf{T/I}
        & \textbf{Gap}
        & \textbf{All}
        & \textbf{All}
        & \textbf{T/I}
        & \textbf{Gap} \\
        \midrule

        Qwen3-VL-8B \cite{bai2025qwen3}
        & 50.8
        & 69.8/20.0
        & 49.8
        & 24.0
        & 59.7/16.2
        & 43.5
        & 81.7
        & 88.6/76.2
        & 12.3
        & 54.4
        & 93.4
        & 98.1/87.4
        & 10.7 \\

        MAI-UI-8B \cite{zhou2025mai}
        & \second{62.9}
        & 78.3/37.9
        & 40.4
        & \second{39.6}
        & 76.5/31.6
        & 44.9
        & \best{88.5}
        & 94.7/83.5
        & 11.2
        & 62.1
        & \second{95.4}
        & 98.2/91.7
        & 6.5 \\

        Qwen3-VL-32B \cite{bai2025qwen3}
        & 57.8
        & 76.0/28.3
        & 47.7
        & 35.0
        & 67.6/27.9
        & 39.6
        & \second{87.3}
        & 91.8/83.7
        & 8.1
        & \second{65.6}
        & 95.0
        & 97.8/91.3
        & \second{6.4} \\

        Qwen3.5-4B \cite{qwen35blog}
        & 49.7
        & 64.8/25.3
        & 39.5
        & 33.2
        & 64.5/26.3
        & 38.2
        & 81.7
        & 87.0/77.4
        & 9.6
        & 56.2
        & 91.8
        & 96.2/86.1
        & 10.1 \\

        Qwen3.5-9B \cite{qwen35blog}
        & 60.7
        & 74.5/38.2
        & \second{36.3}
        & 38.6
        & 63.8/33.1
        & \second{30.7}
        & 84.0
        & 87.8/80.9
        & \second{6.8}
        & 62.4
        & 93.5
        & 97.2/88.6
        & 8.6 \\

        Qwen3.5-27B \cite{qwen35blog}
        & \best{70.1}
        & 82.4/50.2
        & \best{32.2}
        & \best{53.6}
        & 72.2/49.5
        & \best{22.7}
        & \best{88.5}
        & 91.9/85.7
        & \best{6.3}
        & \best{69.5}
        & \best{95.5}
        & 97.6/92.8
        & \best{4.9} \\

        \bottomrule
    \end{tabular}
    }
    \vspace{-1.5em}
\end{table*}

\subsection{Repeated Sampling and Candidate Selection}

The preceding visual and instruction experiments use a single prediction for each sample.
To examine whether additional test-time sampling improves GUI grounding, we keep the screenshot and instruction fixed while varying coordinate generation and candidate selection.
As shown in Figure~\ref{fig:framework} (c), we first obtain a baseline prediction $\hat{\mathbf{p}}_{\mathrm{greedy}}$ using greedy decoding with temperature $0$.
We then independently sample $K=8$ coordinates at temperature $T=0.7$, yielding the candidate set
$\mathcal{P}=\{\mathbf{p}_i\}_{i=1}^{K}$.

To evaluate verifier-free candidate selection, we choose the coordinate medoid, defined as the sampled point with the smallest total Euclidean distance to the remaining candidates $\mathbf{p}_{\mathrm{medoid}}
= \arg\min_{\mathbf{p}_i\in\mathcal{P}} \sum_{\substack{j=1 \\ j\neq i}}^{K} \left\| \mathbf{p}_i-\mathbf{p}_j \right\|_2$.
Unlike coordinate averaging, the medoid is always one of the sampled predictions and does not introduce an interpolated coordinate.
Candidate coverage is measured by Oracle@$K$.
For each sample, we define the corresponding indicator as $o_K=\mathbb{I}\left[
\exists\,\mathbf{p}_i\in\mathcal{P}\ \text{such that}\ \mathbf{p}_i\in B \right]$, where $B$ denotes the ground-truth bounding box.
Oracle@$K$ is obtained by averaging $o_K$ over the evaluation set.
At the dataset level, the differences between Oracle@$K$ and Greedy and between Medoid and Greedy quantify the potential gain from candidate coverage and verifier-free spatial consensus, respectively.
The gap between Oracle@$K$ and Medoid reflects cases in which a correct candidate has been generated but is not reliably selected.

Finally, we quantify sampling dispersion using the normalized mean pairwise distance $\sigma=\frac{2}{K(K-1)} \sum_{i<j} d_{\mathrm{norm}} \left( \mathbf{p}_i, \mathbf{p}_j \right)$, where $d_{\mathrm{norm}}$ is the Euclidean distance after normalizing the horizontal and vertical coordinates by the screenshot width and height.
A larger $\sigma$ indicates greater variation among the sampled coordinates and is used as a test-time proxy for grounding uncertainty.

\section{Experiments}

\begin{figure*}
    \centering
    \includegraphics[width=0.8\linewidth]{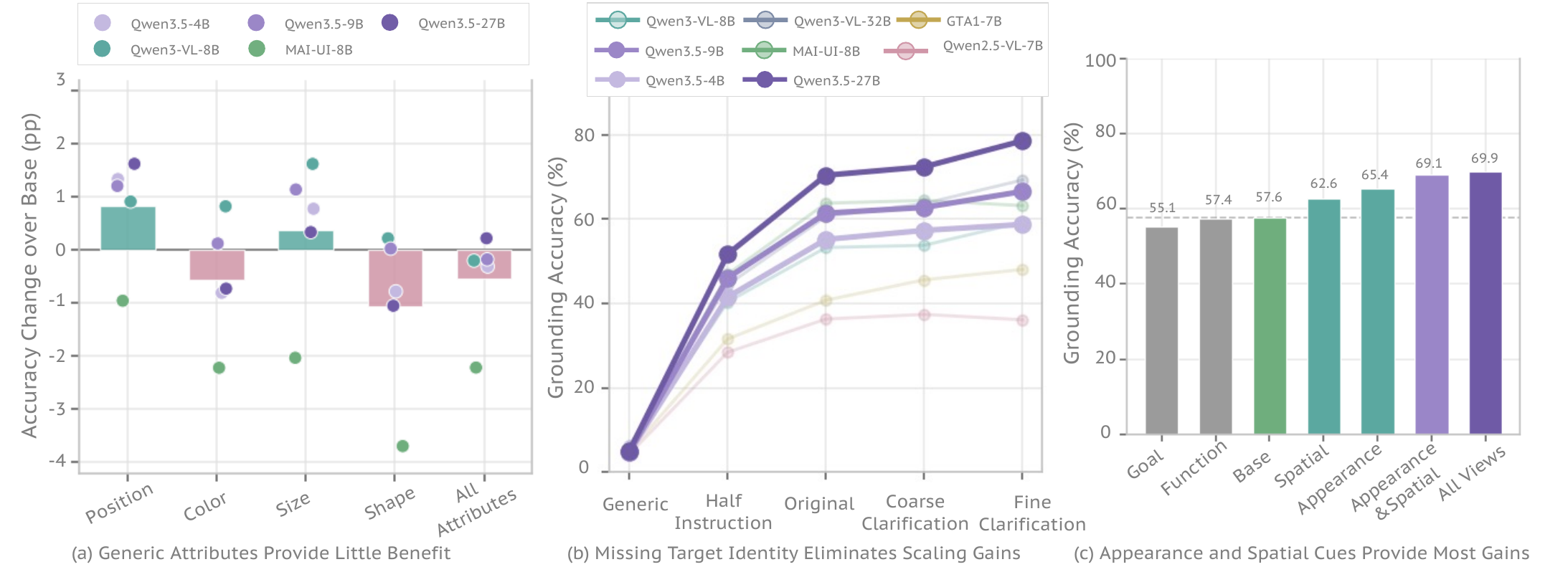}
    \vspace{-1.em}
\caption{
Instruction utility depends on target-identifying information.
\textbf{(a)} Effects of appending annotation-derived coarse attributes, including position, color, size, shape, and their combination.
Bars show the mean accuracy change relative to the original instruction over models and four benchmarks, while points show the corresponding per-model averages.
\textbf{(b)} Grounding accuracy along an information ladder, from a generic instruction without target identity, through truncated and original instructions, to coarse and fine location clarifications.
Qwen3.5 models are emphasized with darker curves.
\textbf{(c)} Grounding accuracy under the original instruction and selected GPT-5.2-generated single-view and multi-view descriptions.
Each bar is averaged over seven models and five benchmarks, and the dashed line marks the original-instruction baseline.
}
    \vspace{-2.0em}
    \label{fig:instruction}
\end{figure*}

\subsection{Experimental Setup}

\noindent\textbf{Environment and Model Configuration.}
All models are deployed on NVIDIA H100 GPUs using vLLM. 
We use model-specific image preprocessing, coordinate spaces, output parsers, and generation limits, and map all predictions back to the original screenshot coordinates. 
The generation limits are sufficient to prevent output truncation. 
Unless otherwise specified, we use greedy decoding with temperature $0$ and leave top-$k$ and top-$p$ unspecified under the vLLM defaults. For repeated sampling, we set the temperature to $0.7$.

\noindent\textbf{Datasets and Models.}
We evaluate on five GUI grounding benchmarks including ScreenSpot-Pro (SS-Pro) \cite{li2025screenspot}, UI-Vision \cite{nayak2025ui}, MMBench-GUI \cite{wang2025mmbench}, OSWorld-G \cite{xie2026scaling}, and ScreenSpot-v2 \cite{screenspotv2-cheng2024seeclick}. 
Full test sets are used for baseline evaluation and multi-view rewriting, and all other experiments use fixed-seed subsets for consistent cross-model comparisons.
We evaluate eight models including Qwen3-VL-8B and 32B \cite{bai2025qwen3}, Qwen3.5-4B, 9B, and 27B \cite{qwen35blog}, Qwen2.5-VL-7B \cite{bai2025qwen25vltechnicalreport}, MAI-UI-8B \cite{zhou2025mai}, and GTA1-7B \cite{yang2026gta1}. 
These models cover general-purpose VLMs, GUI-specific grounders, and multiple scales within the same model family.

\noindent\textbf{Metrics.}
Beyond grounding accuracy, we also measure the gain of the best crop over the full screenshot, the performance drop under irrelevant context, and the gain from fine-grained clarification. 
For repeated sampling, we report Greedy, Medoid, Oracle@$K$, and the normalized mean pairwise distance among sampled coordinates.

\begin{table*}[!t]
    \centering
    \scriptsize
    \setlength{\tabcolsep}{7.5pt}
    \renewcommand{\arraystretch}{0.86}
    \caption{
    Grounding accuracy under different combinations of instruction views.
    Checkmarks indicate the views included in each instruction.
    Best and second-best results in each model column are highlighted in
    \besttext{green} and \secondtext{purple}, respectively.
    }
    \label{tab:multiview_instruction_combinations}
    \vspace{-0.5em}
    \resizebox{0.98\textwidth}{!}{
    \begin{tabular}{l*{4}{c}*{7}{c}}
        \toprule
        \textbf{Setting}
        & \textbf{App}
        & \textbf{Spa}
        & \textbf{Fun}
        & \textbf{Goa}
        & \textbf{Qwen3-VL-8B}
        & \textbf{MAI-UI-8B}
        & \textbf{Qwen2.5-VL-7B}
        & \textbf{GTA1-7B}
        & \textbf{Qwen3.5-4B}
        & \textbf{Qwen3.5-9B}
        & \textbf{Qwen3.5-27B} \\
        \midrule

        Base
        &  &  &  &
        & 60.8 & 66.0 & 32.8 & 37.9 & 62.5 & 67.8 & 75.3 \\

        \midrule

        Single
        & \checkmark &  &  &
        & 70.8 & 70.6 & 37.4 & 49.8 & 74.0 & 76.1 & 78.9 \\

        Single
        &  & \checkmark &  &
        & 66.4 & 68.3 & 34.8 & 50.6 & 69.4 & 71.9 & 76.5 \\

        Single
        &  &  & \checkmark &
        & 62.1 & 64.3 & 32.7 & 40.2 & 63.1 & 67.4 & 71.7 \\

        Single
        &  &  &  & \checkmark
        & 57.8 & 65.4 & 31.6 & 33.7 & 60.9 & 64.8 & 71.7 \\

        \midrule

        Pair
        & \checkmark & \checkmark &  &
        & 74.2 & 74.4 & 38.8 & 55.9 & 77.9 & 79.6 & 82.9 \\

        Pair
        & \checkmark &  & \checkmark &
        & 71.7 & 71.4 & 37.5 & 54.9 & 73.9 & 77.6 & 80.2 \\

        Pair
        & \checkmark &  &  & \checkmark
        & 72.9 & 72.7 & 38.1 & 54.7 & 75.1 & 78.8 & 81.5 \\

        Pair
        &  & \checkmark & \checkmark &
        & 68.9 & 71.5 & 36.2 & 52.8 & 72.6 & 76.1 & 80.4 \\

        Pair
        &  & \checkmark &  & \checkmark
        & 68.8 & 72.0 & 36.5 & 52.5 & 73.3 & 76.3 & 80.5 \\

        Pair
        &  &  & \checkmark & \checkmark
        & 62.5 & 67.1 & 33.7 & 37.5 & 64.4 & 68.2 & 74.1 \\

        \midrule

        Triple
        & \checkmark & \checkmark & \checkmark &
        & \second{75.0}
        & 74.9
        & 38.9
        & \best{57.1}
        & 77.9
        & \second{80.7}
        & 83.6 \\

        Triple
        & \checkmark & \checkmark &  & \checkmark
        & 74.9
        & \second{75.1}
        & \second{39.1}
        & \second{56.9}
        & \best{78.1}
        & \second{80.7}
        & \second{84.1} \\

        Triple
        & \checkmark &  & \checkmark & \checkmark
        & 72.9 & 72.9 & 38.1 & 55.0 & 75.0 & 78.7 & 81.3 \\

        Triple
        &  & \checkmark & \checkmark & \checkmark
        & 70.1 & 72.4 & 36.7 & 52.9 & 73.1 & 76.8 & 81.0 \\

        \midrule

        All
        & \checkmark & \checkmark & \checkmark & \checkmark
        & \best{75.6}
        & \best{75.2}
        & \best{39.2}
        & 55.7
        & \second{78.0}
        & \best{81.0}
        & \best{84.5} \\

        \bottomrule
    \end{tabular}
    }
    \vspace{-2.0em}
\end{table*}

\subsection{Visual Scaling Is Governed by Target Occupancy}
Table~\ref{tab:baseline_type_breakdown} shows that visual difficulty is concentrated in SS-Pro and UI-Vision, where icon targets are substantially harder than text targets. 
For several models, the text--icon gap reaches 39--50 points, while the gap is much smaller on MMBench-GUI and ScreenSpot-v2. 
Within the Qwen3.5 family, this gap narrows from 4B to 27B but remains substantial, indicating that model scaling alone does not eliminate errors on small targets.
Figure~\ref{fig:vis_scaling_bar} (a) shows that target-centered cropping has a nonmonotonic effect. 
Accuracy generally improves as the crop becomes smaller, peaks when 6.25\%--25\% of the original screenshot area is retained, and then declines as more surrounding context is removed.
Figure~\ref{fig:vis_scaling_bar} (b) further shows that the largest gains occur on SS-Pro and UI-Vision, which contain more small targets, whereas ScreenSpot-v2 benefits little. 
The best crop ratio also varies across benchmarks and models, showing that visual scaling cannot be determined by a single crop ratio.
 
Figures~\ref{fig:visual_gated} (a) and (b) reveal a more consistent relationship with target occupancy. 
Targets with low full-screen OCC benefit most from cropping, whereas targets with full-screen OCC above 2\% are generally harmed. 
After cropping, gains remain broadly positive below approximately 2\% OCC, weaken between 2\% and 5\%, and become negative beyond roughly 5\%. 
Among full-screen errors that are rescued by cropping, the median OCC at the first successful crop is only 0.32\%.
These results show that visual scaling requires only modest target enlargement and quickly saturates once the target occupies a sufficient fraction of the input.
Figure~\ref{fig:visual_gated} (c) separately examines surrounding context. Adding irrelevant interface content while preserving the target screenshot and its scale reduces accuracy for most models. 
However, performance on small targets does not consistently predict robustness to irrelevant context, and neither effect scales monotonically with model size. Target scale and surrounding context therefore affect grounding differently and should not be conflated.

\begin{findingbox}
\textcolor{tpurple}{\textbf{Finding 1.}}
Visual scaling is governed by target occupancy rather than by a fixed crop ratio. 
Cropping primarily helps targets with low occupancy, requires only modest enlargement to recover many errors, and becomes harmful once further cropping removes useful surrounding context. 
% Target scale and irrelevant context have distinct effects, neither of which is consistently resolved by increasing model size.
\end{findingbox}

\begin{figure*}
    \centering
    \includegraphics[width=0.85\linewidth]{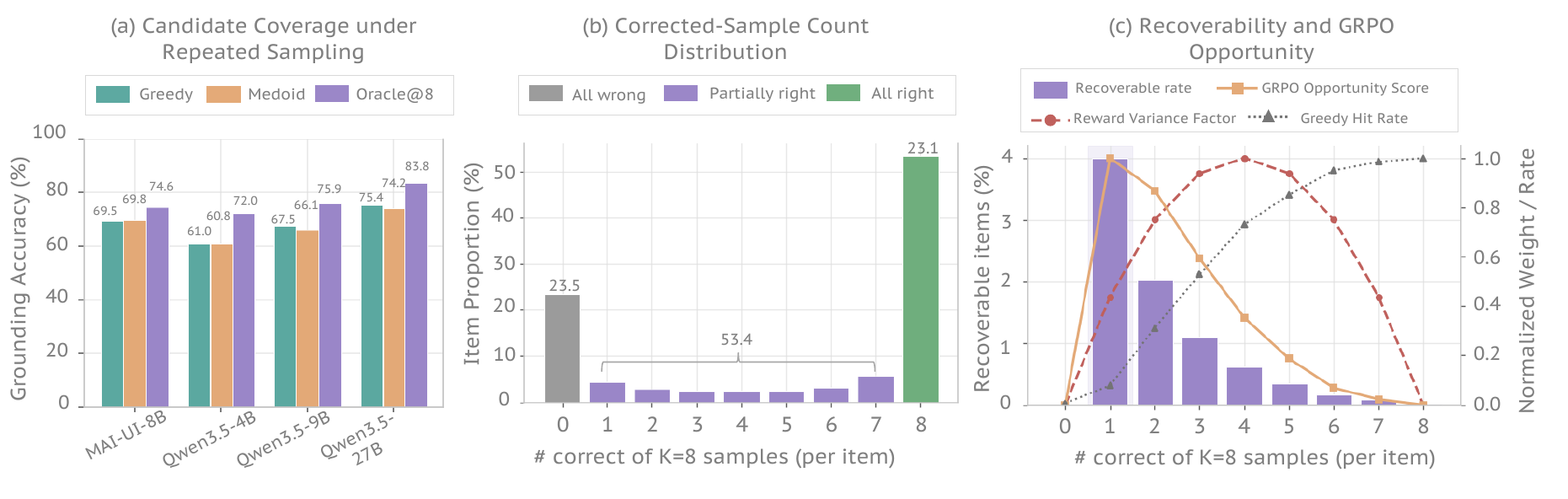}
    \vspace{-1.em}
\caption{
Repeated sampling improves candidate coverage but leaves a selection gap.
Sampling uses $K=8$ predictions at $T=0.7$, while Greedy uses one prediction at $T=0$.
\textbf{(a)} Grounding accuracy under Greedy, coordinate-medoid selection, and Oracle@$8$, averaged over five benchmarks with 500 items each.
Medoid selects the coordinate with the smallest total distance to the other samples, whereas Oracle@$8$ counts an item as correct if any sample hits the target.
\textbf{(b)} Distribution of the number of correct samples $m$, pooled over four models and five benchmarks ($N=10{,}000$), with items categorized as all wrong ($m=0$), partially correct ($0<m<8$), or all correct ($m=8$).
\textbf{(c)} Recoverable items as a function of $m$.
Bars show the proportion of items missed by Greedy but covered by Oracle@$8$; curves show the normalized GRPO opportunity score, reward-variance factor, and Greedy hit rate.
}
    \vspace{-1.5em}
    \label{fig:sampling}
\end{figure*}

% \vspace{-2.5em}
\subsection{Instruction Scaling Depends on Target Identification}
% \vspace{-1.5em}

Figure~\ref{fig:instruction} (a) shows that adding coarse target attributes provides little consistent benefit. 
Position yields only a small average improvement, while color, size, and shape are neutral or harmful for several models. 
Even target descriptions derived directly from ground truth therefore do not necessarily improve grounding.
Figure~\ref{fig:instruction} (b) a much stronger dependence on target-identifying information. Removing the information needed to identify the referred element reduces all evaluated models to approximately 4\%--6\% accuracy and nearly eliminates the performance gap between Qwen3.5 models of different sizes. 
Restoring the original instruction recovers most of the lost performance, while additional location information provides smaller gains. 
Increasing model size therefore improves the use of available instruction information but does not compensate when target-identifying information is missing.

Figure~\ref{fig:instruction} (c) and Table~\ref{tab:multiview_instruction_combinations} further show that different additional target descriptions contribute very differently. Appearance descriptions improve average accuracy from 57.6\% to 65.4\%, spatial descriptions reach 62.6\%, and combining the two reaches 69.1\%. 
In contrast, function descriptions provide little improvement and goal descriptions reduce average accuracy. 
Combining all four description types reaches 69.9\%, only 0.8 points above appearance and spatial descriptions alone. 
Thus, most of the gain comes from additional descriptions that make the target easier to distinguish by its appearance and location, rather than from adding more target-related information in general.

\begin{findingbox}
\textcolor{tpurple}{\textbf{Finding 2.}}
Instruction scaling depends on target-identifying information, not more description. Larger models cannot compensate when this information is missing, while appearance and spatial descriptions provide most of the gain.
\end{findingbox}

\subsection{Sampling Improves Coverage but Not Selection}

Figure~\ref{fig:sampling} (a) suggests that repeated sampling improves candidate coverage but not candidate selection. 
Compared with greedy decoding, Oracle@$8$ improves average grounding accuracy by 5.1--11.0 points across the four evaluated models, showing that additional samples often contain a correct coordinate when the greedy prediction is wrong. 
In contrast, coordinate-medoid selection changes accuracy by only $-1.4$ to $+0.3$ points, leaving most of this gain unrealized.
Figure~\ref{fig:sampling} (b) shows where this gap arises. 
Across the pooled model--item pairs, 23.1\% contain both correct and incorrect samples, meaning that a correct candidate is generated but is not consistently selected. 
The prevalence of these cases varies substantially across models, from 32.6\% for Qwen3.5-4B to 9.2\% for MAI-UI-8B, indicating that the potential benefit of additional sampling is strongly model-dependent.

Figure~\ref{fig:sampling} (c) further explains why geometric selection captures little of this potential gain. 
Among items rescued by Oracle@$8$, correct coordinates often appear only once or twice among eight samples and therefore remain rare among the generated candidates. 
A medoid favors the dominant spatial cluster, making these rare correct candidates unlikely to be selected. 
As correct candidates become more frequent, greedy decoding is also increasingly likely to be correct, leaving less additional gain for repeated sampling.

\begin{findingbox}
\textcolor{tpurple}{\textbf{Finding 3.}}
Prediction sampling improves candidate generation but not candidate selection. 
Additional samples frequently contain correct coordinates missed by greedy decoding, but these coordinates are often rare among the sampled candidates and are therefore not recovered by geometric selection.
\end{findingbox}

\section{Conclusion}

In this work, we propose a systematic framework for studying test-time scaling in GUI grounding across visual input, instruction information, and output sampling. 
Our results show that scaling is effective only when it addresses the active bottleneck. 
Visual scaling improves target visibility, instruction scaling resolves target ambiguity, and repeated sampling expands candidate coverage but still requires reliable selection. 
These findings motivate adaptive scaling based on target visibility, identity, and selection uncertainty.

\clearpage
\bibliographystyle{IEEEtran}
\bibliography{IEEEfull}

\end{document}